\documentclass[a4paper,fleqn]{cas-dc}

\usepackage[numbers]{natbib}

\def\tsc#1{\csdef{#1}{\textsc{\lowercase{#1}}\xspace}}
\tsc{WGM}
\tsc{QE}
\tsc{EP}
\tsc{PMS}
\tsc{BEC}
\tsc{DE}

\begin{document}
\let\WriteBookmarks\relax
\def\floatpagepagefraction{1}
\def\textpagefraction{.001}
\shorttitle{fMRI-Diffusion for MDD Diagnosis}
\shortauthors{M.A. Hasan et~al.}

\title [mode = title]{fMRI-Diffusion: Generating fMRI Time Series Via a Temporal Transformer Diffusion Model for Major Depressive Disorder Diagnosis}                      



\author[1]{Muhammad Asif Hasan}[orcid=0000-0003-3133-9106]
\ead{muhammadasif.hasan@griffithuni.edu.au}

\author[1]{Yanming Zhu}[orcid=0000-0002-8238-8090]
\ead{yanming.zhu@griffith.edu.au}

\author[1]{Xuefei Yin}[orcid=0000-0002-5784-7419]
\ead{x.yin@griffith.edu.au}

\author[1]{Alan Wee-Chung Liew}[orcid=0000-0001-6718-7584]
\cormark[1]
\ead{a.liew@griffith.edu.au}

\affiliation[1]{organization={School of Information and Communication Technology, Griffith University},
                city={Gold Coast},
                state={QLD},
                postcode={4215},
                country={Australia}}

\cortext[cor1]{Corresponding author}

\begin{abstract}
\textbf{Background:} Diagnosing Major Depressive Disorder (MDD) from functional magnetic resonance imaging (fMRI) using functional connectivity (FC) analysis requires large amounts of labeled data that are scarce in clinical settings. Existing augmentation methods synthesize FC matrices, which compress fMRI recordings into static pairwise summaries and discard the temporal structure of the original time series.
 
\textbf{Methods:} We propose fMRI-Diffusion, a framework that directly synthesizes region-of-interest (ROI) level fMRI time series rather than FC matrices. A Temporal Transformer serves as the denoising network within a denoising diffusion probabilistic model (DDPM), treating each time point as a token to model temporal dependencies through self-attention. A supervised pretraining strategy initializes the Transformer with task-relevant representations before diffusion training. The synthesized time series are subsequently used to compute FC matrices for downstream classification.
 
\textbf{Results:} Experiments on the REST-meta-MDD dataset show that augmenting training data with our synthetic fMRI time series consistently improves diagnostic accuracy across ten classification models, six brain parcellation atlases, and three independent acquisition sites. The proposed method outperforms five state-of-the-art FC-based synthesis methods in MDD classification, with accuracy improvements of up to 3.7 percentage points over the strongest baseline. Ablation experiments confirm the contributions of both the Transformer-based denoiser and the supervised pretraining strategy. Distributional fidelity metrics, including KL divergence, Wasserstein distance, and the Kolmogorov-Smirnov statistic, remain below 0.06 across all evaluated conditions, indicating close agreement between real and synthetic distributions.
 
\textbf{Conclusion:} Synthesizing fMRI time series before FC computation preserves temporal information lost in matrix-level augmentation. The proposed framework offers a practical augmentation strategy for MDD diagnosis under limited data and may extend to other data-scarce neuroimaging tasks.
\end{abstract}




\maketitle

\section{Introduction}

Major Depressive Disorder (MDD) is a prevalent and debilitating mental illness, and achieving an accurate diagnosis remains a major clinical challenge \cite{Vickers2024, Zhou2024, Qaseem2023}. Neuroimaging techniques, particularly functional magnetic resonance imaging (fMRI), have emerged as promising tools for providing objective biomarkers of MDD. fMRI captures blood-oxygen-level-dependent (BOLD) signal fluctuations, enabling the analysis of brain functional connectivity (FC) patterns that reflect neural interactions underlying emotional and cognitive processes \cite{Leskinen2024, Kim2022}. By identifying FC abnormalities associated with MDD, fMRI-based FC analysis offers the potential for more objective, reliable, and early diagnosis \cite{Ji2021}.

The advancements in deep learning (DL) have facilitated the development of various DL-based approaches for analyzing FC in MDD diagnosis \cite{Ji2022, Zhu2023}. However, a major challenge of these methods lies in their dependence on large-scale datasets, which are often limited in clinical settings \cite{Tan2024}. When brain topological representations are constructed using limited FC data, critical information essential for accurate diagnosis may be overlooked, leading to unreliable diagnosis \cite{Lu2021}. Therefore, effective and reliable data augmentation methods are needed to enhance the robustness of FC analysis and improve diagnosis.

In recent years, several data synthesis methods have been proposed. For example, Oh et al. \cite{oh2023graph} utilized generative adversarial networks (GANs) to synthesize FC data, enhancing the classification of MDD. Similarly, Tan et al. \cite{Tan2024} introduced a deep convolutional GAN (DCGAN)-based framework to augment FC data. However, these methods primarily synthesize FC matrices or FC-derived graphs, which are dominated by pairwise relationships. Because FC is typically computed from pairwise correlations, it compresses high-dimensional fMRI signals into simplified connectivity summaries and may discard multivariate temporal structure and non-linear dependencies~\cite{Varley2023,Li2023}. Considering fMRI data contains rich and dynamic spatial-temporal features, a more effective strategy would be to directly synthesize fMRI time series rather than generating FC matrices. However, for MDD, most augmentation pipelines still operate on precomputed FC or FC-derived graphs, and synthesis of direct rs-fMRI time series remains underexplored.

To fill this research gap, we propose fMRI-Diffusion, a diffusion-based framework designed to synthesize rs-fMRI time-series data for MDD data augmentation to improve the accuracy of MDD diagnosis. Specifically, we adapt a Transformer encoder architecture as the denoising backbone, incorporating temporal positional encoding to model dynamic dependencies across rs-fMRI time points. The Temporal Transformer serves as the denoising backbone within a denoising diffusion probabilistic model (DDPM), guiding iterative denoising to generate temporally coherent, high-fidelity time series data. Furthermore, we introduce a supervised pretraining strategy to improve representation quality and training stability under limited data conditions. Extensive experiments demonstrate that fMRI-Diffusion produces realistic and biologically plausible fMRI time series data and improves downstream MDD diagnostic performance compared with state-of-the-art FC-based synthesis methods. The main contributions of this paper are summarized as follows:
\begin{itemize}
    \item We formulate MDD-oriented data augmentation as region of interest (ROI) time-series synthesis rather than FC-matrix synthesis. By generating ROI sequences prior to FC computation, our approach preserves ROI-level temporal dynamics and higher-order temporal interactions that are otherwise collapsed by correlation-based FC representations.
    \item We propose fMRI-Diffusion, a generative framework that employs a Transformer-based denoising network within a DDPM to synthesize high-fidelity ROI rs-fMRI time series. The denoising network is designed to operate along the temporal axis of fMRI sequences, modelling dependencies across time points, while the DDPM provides the generative mechanism for structured and realistic synthesis via iterative denoising. Furthermore, we introduce a supervised pretraining strategy for the Temporal Transformer to enhance representation learning and optimization stability under limited data. This framework provides a robust and generalizable solution for fMRI data augmentation under limited data conditions.
    \item Extensive experiments on the Resting-state MDD dataset demonstrate that fMRI-Diffusion significantly outperforms state-of-the-art FC-based methods in downstream diagnostic performance. The superiority of our framework is consistently validated across different brain atlases, multiple MDD datasets, and various classification models, highlighting its robustness, effectiveness, and generalizability in generating high-fidelity fMRI time-series data for improved MDD diagnosis.
\end{itemize}

\section{Related Background}

\subsection{Denoising Diffusion Probabilistic Model}

Denoising Diffusion Probabilistic Models (DDPM) are a class of generative models that learn to synthesize data by gradually reversing a diffusion process that incrementally corrupts a sample with Gaussian noise \cite{Ho2020}. Specifically, DDPM consists of two stages: a forward diffusion process, where noise is progressively added to the original data over multiple time steps, and a reverse denoising process, where a neural network learns to reconstruct clean data from noisy inputs. During the forward process, a clean sample $x_0 \sim P(x_0)$ is gradually transformed into a noisy version $x_T$ by applying Gaussian perturbations at each step, with the level of corruption controlled by a predefined variance schedule. The reverse process is modeled as a series of learned conditional distributions, parameterized by a neural network, which predict the mean and variance required to iteratively denoise the sample. The training of DDPM is formulated as a minimization of a simplified mean-squared error loss, where the neural network is optimized to predict the Gaussian noise added at each diffusion step. By accurately estimating the noise, the network learns to approximate the reverse transitions of the diffusion process, enabling the progressive reconstruction of clean data from noisy inputs. Once trained, the model can generate new data by starting from pure Gaussian noise and applying the learned denoising steps sequentially, producing samples that closely resemble the training distribution.

DDPM offers several advantages that make it well-suited for synthesizing fMRI time-series data. Its stepwise generation process allows fine-grained control over sample structure and fidelity, which is critical for preserving the rich spatial and temporal dynamics of fMRI signals. In addition, the probabilistic nature of DDPM ensures robustness to data scarcity and noise, addressing common challenges in medical imaging. Leveraging these properties, we integrate a Transformer-based denoising network into the DDPM framework to guide the diffusion process and generate high-fidelity, temporally coherent synthetic fMRI time series.

\subsection{Data Synthesis for MDD Diagnosis}

In recent years, various data synthesis techniques, such as Gaussian Mixture Models (GMM), Variational Autoencoders (VAE), and GAN, have been extensively applied to augment limited neuroimaging datasets for brain disorder diagnosis \cite{Qiang2023, Zhuang2019, li2021brainnetgan, Yan2021}. However, data synthesis specifically targeting MDD diagnosis remains relatively unexplored, even though data scarcity continues to be a major challenge. Several recent studies have attempted to address this limitation using GAN-based frameworks. For example, Oh et al. \cite{oh2023graph} designed a graph-convolutional GAN to generate synthetic FC matrices, effectively mitigating mode collapse by modeling the topological structure of brain networks. Similarly, Tan et al. \cite{Tan2024} proposed a deep convolutional GAN (DCGAN) to produce signed-weighted FC representations that preserve spatial relationships. Although these methods have shown improved classification performance, they are inherently limited by their reliance on precomputed FC matrices derived from pairwise correlations of BOLD signals. Such representations oversimplify the high-dimensional and dynamic nature of fMRI data and often overlook critical spatial-temporal dynamics essential for understanding MDD-related brain dysfunctions \cite{Varley2023, Li2023}. Moreover, GANs are prone to mode collapse, which compromises the diversity of synthetic samples and undermines the goal of effective data augmentation \cite{Hou2022}.

To date, several studies have explored generative modeling and augmentation of fMRI time-series data in broader neuroimaging settings. However, diffusion-based synthesis of rs-fMRI time series for MDD diagnosis, particularly in an ROI-parcellated time-series representation, has received limited attention. Working directly with rs-fMRI timepoints is intrinsically challenging. Resting-state fMRI exhibits spontaneous fluctuations, each subject's scan may span multiple brain states, and there is no guaranteed temporal correspondence across subjects. These properties make it difficult to extract consistent temporal patterns across individuals, which has prompted many MDD studies to rely on static or dynamic FC-based summaries. To address this, we propose fMRI-Diffusion, a framework that integrates a Transformer-based denoising network within a DDPM to synthesize ROI-wise rs-fMRI time series. By generating time series prior to FC construction, the proposed method retains temporally resolved ROI dynamics that are otherwise compressed when using correlation-based FC representations, thereby supporting more effective data augmentation for downstream MDD diagnosis.

\section{Methodology}
\begin{figure*}[htbp]
	\centering
	\includegraphics[width=\linewidth]{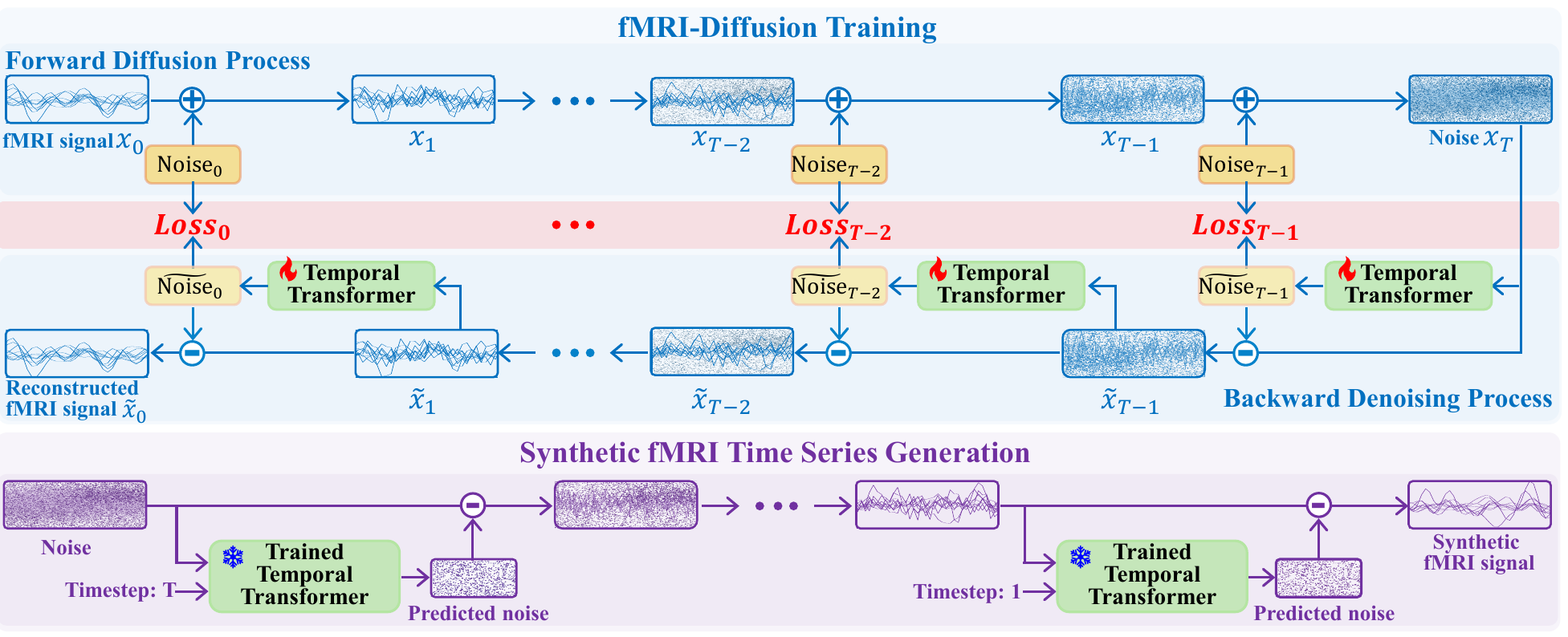}
	\caption{Overview of the proposed fMRI-Diffusion framework for synthesizing fMRI time-series data. The framework consists of two main stages. (1) Training (top), which follows the standard DDPM framework with a forward diffusion process and a reverse denoising process. In the forward process, Gaussian noise is progressively added to the original fMRI time series. In the reverse process, a Temporal Transformer serves as the denoising network to predict and remove noise at each diffusion step, enabling the reconstruction of clean fMRI signals. (2) Generation (bottom), where new fMRI sequences are synthesized by starting from pure noise and iteratively applying the trained Temporal Transformer in the reverse denoising process.}
	\label{fig1}
\end{figure*}
\subsection{Framework of the Proposed fMRI-Diffusion}

The overall workflow of the proposed fMRI-Diffusion framework is illustrated in Fig. \ref{fig1}. It consists of two main phases: model training and data generation. The framework is built upon the DDPM, which learns to synthesize fMRI time-series data through a forward diffusion process and a backward denoising process. During the training phase, Gaussian noise is progressively added to each input fMRI time series $x_0$ over $T$ time steps. At each step $t$, the data $x_t$ is corrupted by adding $\text{Noise}_t$. After $T$ steps, this process results in a fully corrupted signal $x_T$ that becomes indistinguishable from pure noise. In the backward denoising process, a Temporal Transformer is employed as the denoising network to predict the noise component $\tilde{\text{Noise}}_t$ at each time step $t$. By minimizing the discrepancy between the predicted and true noise, the Temporal Transformer learns to iteratively remove noise and reconstruct high-fidelity fMRI time series. After training, synthetic fMRI data are generated by starting from pure Gaussian noise and applying the Temporal Transformer through the learned reverse diffusion process. This procedure enables the model to produce realistic fMRI time-series signals that preserve the rich temporal dynamics inherent in the original data.

\subsection{Denoising Diffusion Probabilistic Modeling for fMRI}

We employ DDPM as the generative backbone for synthesizing fMRI time-series data. Unlike conventional applications of DDPM in image domains, our formulation is specifically adapted to ROI-parcellated rs-fMRI time-series signals. In this setting, each training sample corresponds to a time series rather than a 2D or 3D spatial structure. The forward diffusion process gradually perturbs a clean fMRI signal $x_0$ by injecting Gaussian noise over $T$ time steps, producing a noisy sample $x_t$ at each step and eventually a fully corrupted signal $x_T$. This is defined as:
\begin{equation}
    q(x_t|x_{t-1}) = \mathcal{N}(\sqrt{1-\beta_t}\, x_{t-1}, \, \beta_t I),
\end{equation}
where $\beta_t$ controls the noise schedule at step $t$, and $I$ is the identity matrix.

The reverse process learns to iteratively remove noise and reconstruct the clean signal $x_0$, which is modeled as:
\begin{equation}
    p_\theta(x_{t-1}|x_t) = \mathcal{N}(\mu_\theta(x_t, t), \, \Sigma_\theta(x_t, t)),
\end{equation}
where $\theta$ denotes the parameters of the neural network. The network predicts the added noise $\epsilon_\theta(x_t, t)$ at each step, which is then used to compute the mean $\mu_\theta(x_t, t)$ of the reverse Gaussian transition. The variance term $\Sigma_\theta(x_t, t)$ represents the uncertainty in the reverse step and follows a fixed schedule, as in the original DDPM formulation. The model is trained by minimizing the noise-prediction loss:
\begin{equation}
L = \mathbb{E}_{x_0, \epsilon, t}\!\left[\|\epsilon - \epsilon_\theta(x_t, t)\|^2\right],
\end{equation}
where $\epsilon$ is the true Gaussian noise added in the forward process.

By applying DDPM to fMRI signals, we leverage its fine-grained generative control and stable training dynamics to model the rich temporal patterns inherent in BOLD time series. Once trained, the model synthesizes realistic fMRI sequences by starting from Gaussian noise and progressively denoising them.

\subsection{Temporal Transformer Denoising Network}

\begin{figure*}[htbp]
	\centering
	\includegraphics[width=\linewidth]{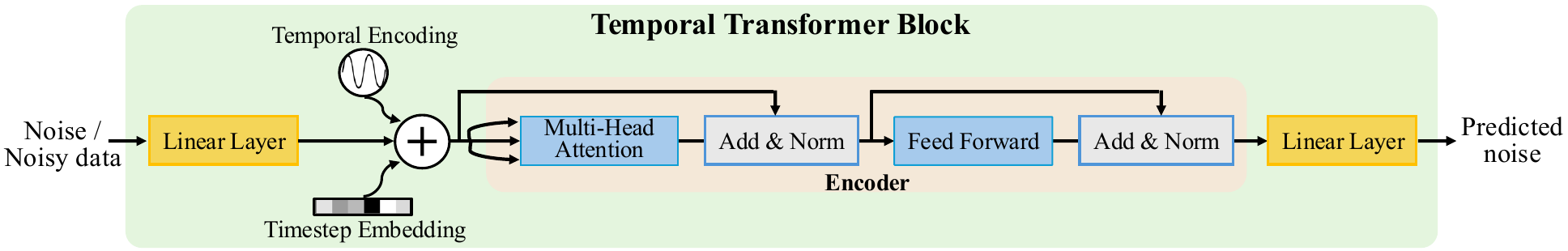}
	\caption{Architecture of the Transformer-based denoising network used in the proposed fMRI-Diffusion framework. The model combines temporal encoding and timestep embedding with the noisy fMRI input, processes the sequence through multi-head self-attention and feed-forward layers, and outputs the predicted noise for each diffusion step to guide the reverse denoising process.}
	\label{fig2}
\end{figure*}

To model the temporal dependencies in fMRI time-series data, we employ a Transformer encoder as the denoising network within the DDPM framework (Fig. \ref{fig2}). The architecture follows the standard Transformer encoder design~\cite{vaswani2017attention} and draws on recent work that has demonstrated the effectiveness of Transformer-based backbones for diffusion models~\cite{peebles2023scalable}. However, unlike prior applications that operate on spatial image patches, our formulation treats each time point of the ROI-parcellated fMRI signal as a token in the input sequence. This temporal formulation allows the self-attention mechanism to capture dependencies across the time axis of the fMRI signal, which is the relevant dimension for modelling BOLD signal dynamics.

The network first projects the noisy input into a latent feature space using a linear layer, followed by the addition of two key embeddings. The \textit{temporal encoding} provides sequence-order information along the time axis using a fixed sinusoidal encoding scheme, as in the Transformer architecture, defined as $\text{TE}_{(t, 2i)} = \sin(t / 10000^{2i/D})$ and $\text{TE}_{(t, 2i+1)} = \cos(t / 10000^{2i/D})$, where $t$ denotes the temporal index, $i$ is the feature dimension, and $D$ is the hidden dimension size. The \textit{timestep embedding} encodes the diffusion step index $t$ and is implemented as a learnable vector obtained through a small MLP. These embeddings together enable the model to capture both the intrinsic temporal structure of the fMRI signal and the generative progression of the diffusion process.

The core of the model is a Transformer encoder composed of a multi-head self-attention layer and a feed-forward sublayer, each followed by residual connections and layer normalization:
\begin{equation}
\text{Attention}(Q,K,V) = \text{softmax}\left(\frac{QK^{\top}}{\sqrt{d_k}}\right)V,
\end{equation}
where $Q$, $K$, and $V$ are linear projections of the input sequence. The attention mechanism enables the model to learn long-range dependencies among temporal points, effectively capturing slow-varying neural activity patterns. The subsequent feed-forward layer refines these representations and enhances nonlinear temporal feature interactions. A final linear projection maps the encoded features to the predicted noise.

This configuration allows the denoising network to capture both short- and long-range temporal relationships in fMRI signals. By operating on the full temporal sequence rather than on precomputed pairwise summaries, the model has access to richer temporal structure during the denoising process, which supports the generation of temporally coherent fMRI time-series data.

\subsection{Pretraining Strategy and Model Optimization}
\label{sec:pretrain}

Training diffusion models on high-dimensional fMRI time series can be difficult in settings with limited sample sizes and structured temporal dependencies. To support effective optimization in this process, we adopt a two-stage training procedure that provides an informed initialization for the denoising network. Recent studies have shown that supervised and denoising objectives can be productively combined in multi-stage training pipelines to improve representation quality under limited labeled data~\cite{brempong2022denoising}, and that diffusion-trained networks can learn features that are beneficial for discriminative tasks~\cite{xiang2023denoising}. These findings suggest that classification and denoising representations share useful structure, motivating the transfer of learned features between the two objectives. However, the use of supervised classification pretraining to initialize the denoising backbone of a diffusion model for data synthesis has not, to our knowledge, been previously explored. Accordingly, we pretrain the Temporal Transformer on a supervised objective before diffusion training. This stage encourages the model to learn representations from real fMRI sequences, providing a suitable initialization and promoting stable optimization during subsequent diffusion training.

Specifically, the Temporal Transformer is first pretrained on a supervised classification task using the original fMRI time series. The objective is to distinguish between MDD and healthy control samples, encouraging the model to learn meaningful latent features that reflect underlying neural activity patterns. During pretraining, hyperparameters are selected via a grid search using 5-fold cross-validation on the training data only within each outer fold to avoid using information from the held-out test fold and to mitigate overfitting. The learned weights of the Transformer encoder are then transferred to initialize the denoising network within the diffusion framework, allowing the model to leverage prior knowledge of temporal dynamics when modeling the reverse diffusion process.

In contrast to conventional augmentation methods that directly synthesize precomputed FC matrices or graph features, our method operates in the ROI time-series space and does not require FC computation during generation. We use standard atlas parcellation only to extract ROI signals from preprocessed fMRI volumes, which improves signal-to-noise ratio and reduces dimensionality while retaining temporally resolved dynamics within each ROI. This pretraining paradigm enables the model to learn disorder-relevant temporal representations from ROI sequences, improving initialization and stabilizing diffusion training under limited data.

\section{Experimental Results}

\subsection{Dataset and Preprocessing}

We evaluate our method on the largest publicly available REST-meta-MDD dataset \cite{Chen2022}, released by the Depression Imaging Research Consortium (DIRECT). The consortium comprises 25 research teams from 17 hospitals across China, providing aggregated resting-state fMRI (rs-fMRI) data from patients diagnosed with MDD and demographically matched normal control (NC) subjects. The dataset represents one of the most standardized open-access rs-fMRI collections for MDD research.

In this work, we specifically used preprocessed data from site 20, which is the largest contributing site, collected using a Siemens Tim Trio 3T scanner at Southwest University. This subset includes rs-fMRI scans from 282 MDD patients and 251 NC subjects, offering a robust and balanced sample for model development and evaluation. Each rs-fMRI sequence contains 242 volumes with a repetition time (TR) of 2s and voxel size of 3.4$\times$3.4$\times$3.4 mm$^3$. 

Preprocessing was performed using the Data Processing Assistant for Resting-State fMRI (DPARSF) software \cite{Yan2010}, following the standard pipeline recommended in \cite{Qin2022}. The first ten volumes of each sequence were removed to account for signal equilibration effects, followed by head motion correction, spatial normalization to the MNI152 space, spatial smoothing with a 6\,mm full-width at half-maximum (FWHM) Gaussian kernel, and band-pass filtering within the frequency range of 0.01-0.1\,Hz. 
All fMRI signals were subsequently detrended and z-score normalized to ensure inter-subject comparability.

\subsection{Experimental Setup}

All experiments were implemented in PyTorch and conducted on a workstation equipped with an NVIDIA RTX~4090 GPU and an Intel Xeon processor. The proposed fMRI-Diffusion framework adopts a cosine variance scheduler for the forward diffusion process to provide a smooth noise schedule across diffusion steps. The diffusion model was trained with $T=1000$ diffusion steps following the standard DDPM setting \cite{Ho2020}.

The Temporal Transformer denoiser comprises $6$ encoder layers with a hidden dimension of $256$ and $8$ attention heads. For diffusion model training, we used the Adam optimizer with an initial learning rate of $1\times 10^{-4}$, weight decay of $1\times 10^{-3}$, and a batch size of $16$, and trained for $500$ epochs with $T=1000$ diffusion steps and a cosine noise schedule. The Transformer parameters were initialized from the best checkpoint obtained from the MDD classification pretraining described in Section~\ref{sec:pretrain}.

For data augmentation, synthetic and real fMRI samples were used in equal proportions, corresponding to a 100\% augmentation ratio. To support an unbiased evaluation, we employed a 5-fold cross-validation protocol with subject-level partitioning. For each fold, all training procedures were performed using the four training folds only, and the remaining fold was used exclusively for testing. The Temporal Transformer was pretrained on the training folds for the MDD versus NC classification task, and the resulting weights were then used to initialize the denoising network of the diffusion model, as described in Section~\ref{sec:pretrain}. The Temporal Transformer configuration in the diffusion model follows the architecture used in the pretraining stage to maintain consistency between supervised pretraining and diffusion training.

All experiments were repeated using five random seeds, and results are reported as mean $\pm$ standard deviation. Model performance was evaluated using accuracy (ACC), sensitivity (SEN), specificity (SPEC), and F1-score (F1) to assess diagnostic performance.

\subsection{Comparison with State-of-the-Art Methods}

To comprehensively evaluate the effectiveness of the proposed fMRI-Diffusion framework, we compared its performance with several representative state-of-the-art (SOTA) data synthesis methods previously applied to MDD diagnosis. These include the Semi-Supervised GAN (SSGAN) \cite{Zhao2020}, Wasserstein GAN with Gradient Penalty (WGAN-GP) \cite{li2021brainnetgan}, Auxiliary Classifier GAN (ACGAN) \cite{Yan2021}, Graph Convolutional conditional GAN (GC-GAN) \cite{oh2023graph}, and Deep Convolutional GAN (DCGAN) \cite{Tan2024}. These models were selected as they represent the SOTA paradigms for synthetic data generation in neuroimaging-based MDD studies, covering both fully supervised and conditional GAN frameworks.

For a fair comparison, the number of samples (MDD and NC subjects) and the brain atlas (Harvard-Oxford (HO) and Automated Anatomical Labeling (AAL)) used in our experiments were matched exactly to those adopted in the original studies, as summarized in Table \ref{tab:mdd_comparison}. In line with prior works, Pearson correlation coefficients between ROI-wise time series were computed to construct FC matrices. Additionally, the same classification backbone was employed to ensure full methodological consistency across all experiments.

\begin{table*}[width=\textwidth,cols=7,pos=htbp]
\caption{Performance comparison of MDD diagnosis using data synthesized by the proposed method and SOTA method. Results are evaluated in terms of ACC, SEN, SPEC, and F1, and reported as mean~$\pm$~standard deviation. The best results are highlighted in bold, and the second-best results are underlined.}
\label{tab:mdd_comparison}
\renewcommand{\arraystretch}{1.1}
\centering
\begin{tabular}{lcccccc}
\toprule[1pt]
\textbf{Method} & \textbf{Atlas} & \textbf{Samples (MDD/NC)} & \textbf{ACC (\%)} & \textbf{SEN (\%)} & \textbf{SPEC (\%)} & \textbf{F1 (\%)} \\
\midrule[1pt]

SSGAN \cite{Zhao2020}      & \multirow{5}{*}{HO}  & \multirow{5}{*}{249/228} & 64.62 ± 4.60 & 68.07 ± 8.00 & 60.84 ± 7.20 & 66.62 ± 4.98 \\
WGAN-GP \cite{li2021brainnetgan}      &                      &                          & 64.02 ± 4.50 & 66.54 ± 7.12 & 61.28 ± 6.61 & 65.77 ± 4.77 \\
ACGAN \cite{Yan2021}       &                      &                          & 64.91 ± 4.42 & 67.68 ± 7.56 & \underline{61.90 ± 8.68} & 66.71 ± 4.36 \\
GC-GAN \cite{oh2023graph}       &                      &                          & \underline{65.18 ± 5.08} & \textbf{68.80 ± 8.09} & {61.28 ± 7.53} & \underline{67.23 ± 5.27} \\
{Ours}                     &                      &                          & \textbf{68.85 ± 1.45} & \underline{68.11 ± 4.25} & \textbf{67.96 ± 5.03} & \textbf{69.63 ± 1.99} \\
\cmidrule(lr){1-7}

DCGAN \cite{Tan2024}       & \multirow{2}{*}{AAL} & \multirow{2}{*}{250/227} & 66.86 ± 3.67 & 66.86 ± 3.67 & --           & 66.69 ± 3.72 \\
{Ours}                      &                     &                          & \textbf{68.53 ± 2.53} & \textbf{69.03 ± 2.80} & \textbf{67.96 ± 3.12} & \textbf{68.85 ± 3.52} \\
\bottomrule[1pt]
\end{tabular}
\end{table*}

Table \ref{tab:mdd_comparison} summarizes the quantitative results in terms of ACC, SEN, SPEC, and F1 across both HO and AAL atlases. Overall, the proposed method consistently surpasses all competing approaches in MDD diagnosis, achieving the best performance across nearly all evaluation metrics. Under the HO atlas, our model demonstrates notable improvements, outperforming the strongest baseline (GC-GAN) by 3.7 percentage points in ACC, 6.7 in SPEC, and 2.4 in F1. A similar trend is observed on the AAL atlas, where our approach again achieves clearly superior results across all metrics. These results confirm the robustness and effectiveness of the proposed method in generating diagnostically meaningful synthetic fMRI data.

\subsection{Effectiveness on Diverse MDD Classifiers}

To evaluate the general effectiveness and transferability of the proposed fMRI-Diffusion, we examined whether the synthetic fMRI data generated by our method can enhance the performance of various MDD diagnosis models. Specifically, we considered two groups of representative models covering a broad spectrum of current MDD classification paradigms.

\subsubsection{Specialized MDD diagnosis models}

We first evaluated our method on four SOTA MDD-specific methods, including the Multi-Atlas Fusion (MAF) model \cite{Lee2024}, the Graph Autoencoder-based Fully Connected Neural Network (GAE-FCNN) \cite{Noman2024}, the Graph Neural Network with Modular Attention (GN-NMA) \cite{Si2025}, and the Dynamic Spatio-Temporal Attention Model (DSAM) \cite{thapaliya2025dsam}. To ensure fair comparison, we followed the same configurations, including sample sizes and brain atlases, as reported in their respective studies. 

\subsubsection{Generalizable neural network classifiers}

To further assess the versatility of our synthetic data, we also tested its impact on six general CNN- and GNN-based architectures widely used for brain disorder diagnosis, including ChebyNet \cite{Defferrard2016}, BrainNetCNN \cite{Kawahara2017}, Graph Attention Network (GAT) \cite{Velikovi2017}, Graph Isomorphism Network (GIN) \cite{Xu2018}, Graph Transformer Network (GTN) \cite{Yun2019}, and GraphSAGE \cite{Hamilton2017}. 
All experiments in this group were conducted using the HO atlas for consistency.

Table \ref{tab:merged_performance_corrected} summarizes the comparative results across all classifiers, both with and without incorporating our synthetic data. As observed, using our synthetic data consistently improves performance across all evaluation metrics (ACC, SEN, SPEC, and F1) and across all tested methods. Notably, methods that originally exhibited moderate performance, such as GAE-FCNN and GTN, achieved significant gains in diagnostic accuracy. These results demonstrate that our synthetic fMRI data not only enhances individual model performance but also generalizes effectively across different network architectures and diagnostic paradigms, confirming the robustness and practical utility of our proposed fMRI time-series synthesis framework.

\begin{table*}[width=\textwidth,cols=8,pos=htbp]
\caption{Performance comparison of SOTA MDD diagnosis models with and without using synthetic data generated by our proposed method. Results are evaluated using ACC, SEN, SPEC, and F1 (mean~$\pm$~standard deviation), with the better-performing results highlighted in bold.}
\label{tab:merged_performance_corrected}

\renewcommand{\arraystretch}{1.05}
\setlength{\tabcolsep}{2.5pt}
\small
\centering

\begin{tabular*}{\textwidth}{@{\extracolsep{\fill}} l l c c c c c c @{}}
\toprule[1pt]
\textbf{Method} &
\textbf{Atlas} &
\begin{tabular}[c]{@{}c@{}}\textbf{With/Without}\\\textbf{Our Synth.\ Data}\end{tabular} &
\begin{tabular}[c]{@{}c@{}}\textbf{Samples}\\\textbf{(MDD/NC)}\end{tabular} &
\textbf{ACC (\%)} &
\textbf{SEN (\%)} &
\textbf{SPEC (\%)} &
\textbf{F1 (\%)} \\
\midrule[1pt]

\multirow{6}{*}{MAF \cite{Lee2024}} 
 & \multirow{2}{*}{AAL}      & W/O & \multirow{2}{*}{245/225} & 63.00 $\pm$ 2.05 & 60.66 $\pm$ 4.58 & 64.88 $\pm$ 1.65 & 63.86 $\pm$ 3.55 \\
 &                           & W/  &                          & \textbf{68.85 $\pm$ 2.85} & \textbf{68.23 $\pm$ 3.36} & \textbf{70.84 $\pm$ 3.93} & \textbf{71.90 $\pm$ 2.55} \\
\cmidrule(lr){2-8}
 & \multirow{2}{*}{HO}       & W/O & \multirow{2}{*}{245/225} & 62.75 $\pm$ 3.19 & 61.76 $\pm$ 6.75 & 65.73 $\pm$ 5.25 & 63.52 $\pm$ 4.75 \\
 &                           & W/  &                          & \textbf{68.00 $\pm$ 2.15} & \textbf{67.95 $\pm$ 3.30} & \textbf{66.79 $\pm$ 3.82} & \textbf{68.37 $\pm$ 2.41} \\
\cmidrule(lr){2-8}
 & \multirow{2}{*}{Craddock} & W/O & \multirow{2}{*}{245/225} & 63.85 $\pm$ 4.28 & 62.45 $\pm$ 8.55 & 63.89 $\pm$ 6.56 & 66.88 $\pm$ 5.50 \\
 &                           & W/  &                          & \textbf{68.25 $\pm$ 3.10} & \textbf{68.89 $\pm$ 3.30} & \textbf{70.56 $\pm$ 3.45} & \textbf{71.23 $\pm$ 3.35} \\
\midrule[0.7pt]

\multirow{2}{*}{GAE-FCNN \cite{Noman2024}} 
 & \multirow{2}{*}{AAL} & W/O & \multirow{2}{*}{250/230} & 62.08 $\pm$ 6.54 & 65.85 $\pm$ 8.07 & 61.53 $\pm$ 5.15 & 63.24 $\pm$ 5.25 \\
 &                      & W/  &                          & \textbf{65.54 $\pm$ 3.54} & \textbf{68.90 $\pm$ 3.88} & \textbf{66.06 $\pm$ 4.25} & \textbf{66.38 $\pm$ 4.63} \\
\midrule[0.7pt]

\multirow{4}{*}{GN-NMA \cite{Si2025}} 
 & \multirow{2}{*}{AAL}      & W/O & \multirow{2}{*}{282/251} & 61.88 $\pm$ 5.25 & 62.54 $\pm$ 4.28 & 64.17 $\pm$ 5.30 & 63.00 $\pm$ 4.55 \\
 &                           & W/  &                          & \textbf{64.75 $\pm$ 3.95} & \textbf{65.56 $\pm$ 4.10} & \textbf{66.28 $\pm$ 4.88} & \textbf{67.33 $\pm$ 3.56} \\
\cmidrule(lr){2-8}
 & \multirow{2}{*}{Craddock} & W/O & \multirow{2}{*}{282/251} & 64.88 $\pm$ 4.83 & 68.25 $\pm$ 4.50 & 62.95 $\pm$ 3.85 & 66.48 $\pm$ 3.25 \\
 &                           & W/  &                          & \textbf{67.88 $\pm$ 2.56} & \textbf{72.25 $\pm$ 2.40} & \textbf{65.95 $\pm$ 2.15} & \textbf{68.83 $\pm$ 2.55} \\
\midrule[0.7pt]

\multirow{6}{*}{DSAM \cite{thapaliya2025dsam}} 
 & \multirow{2}{*}{AAL}      & W/O & \multirow{2}{*}{282/251} & 63.15 $\pm$ 3.28 & 64.12 $\pm$ 3.89 & 65.96 $\pm$ 2.89 & 67.99 $\pm$ 3.25 \\
 &                           & W/  &                          & \textbf{65.33 $\pm$ 2.58} & \textbf{65.96 $\pm$ 2.38} & \textbf{68.65 $\pm$ 2.63} & \textbf{69.10 $\pm$ 2.53} \\
\cmidrule(lr){2-8}
 & \multirow{2}{*}{HO}       & W/O & \multirow{2}{*}{282/251} & 62.55 $\pm$ 3.56 & 63.53 $\pm$ 3.95 & 66.99 $\pm$ 4.56 & 67.03 $\pm$ 3.30 \\
 &                           & W/  &                          & \textbf{64.93 $\pm$ 2.89} & \textbf{65.68 $\pm$ 2.63} & \textbf{69.25 $\pm$ 2.86} & \textbf{68.95 $\pm$ 2.89} \\
\cmidrule(lr){2-8}
 & \multirow{2}{*}{Craddock} & W/O & \multirow{2}{*}{282/251} & 63.36 $\pm$ 4.60 & 63.93 $\pm$ 3.90 & 61.15 $\pm$ 4.38 & 61.75 $\pm$ 4.56 \\
 &                           & W/  &                          & \textbf{66.91 $\pm$ 3.96} & \textbf{67.56 $\pm$ 3.15} & \textbf{63.23 $\pm$ 3.20} & \textbf{64.86 $\pm$ 3.16} \\
\midrule[0.7pt]

\multirow{2}{*}{ChebyNet \cite{Defferrard2016}} 
 & \multirow{2}{*}{HO} & W/O & \multirow{2}{*}{282/251} & 63.00 $\pm$ 3.35 & 62.55 $\pm$ 6.50 & 65.89 $\pm$ 5.20 & 64.35 $\pm$ 4.26  \\
 &                      & W/  &                          & \textbf{69.25 $\pm$ 2.15} & \textbf{68.95 $\pm$ 3.30} & \textbf{67.90 $\pm$ 3.82} & \textbf{69.90 $\pm$ 2.41}\\
\midrule[0.7pt]

\multirow{2}{*}{BrainNetCNN \cite{Kawahara2017}} 
 & \multirow{2}{*}{HO} & W/O & \multirow{2}{*}{282/251} & 63.59 $\pm$ 3.86 & 61.32 $\pm$ 3.06 & 62.31 $\pm$ 6.48 & 63.77 $\pm$ 2.96 \\
 &                      & W/  &                          & \textbf{66.68 $\pm$ 2.57} & \textbf{67.49 $\pm$ 4.00} & \textbf{67.62 $\pm$ 4.75} & \textbf{66.29 $\pm$ 2.63} \\
\midrule[0.7pt]

\multirow{2}{*}{GAT \cite{Velikovi2017}}        
 & \multirow{2}{*}{HO} & W/O & \multirow{2}{*}{282/251} & 64.03 $\pm$ 2.36 & 59.19 $\pm$ 3.80 & 66.88 $\pm$ 1.60 & 62.32 $\pm$ 3.30 \\
 &                      & W/  &                          & \textbf{67.87 $\pm$ 4.17} & \textbf{65.51 $\pm$ 7.17} & \textbf{69.97 $\pm$ 5.44} & \textbf{66.18 $\pm$ 4.77} \\
\midrule[0.7pt]

\multirow{2}{*}{GIN \cite{Xu2018}}        
 & \multirow{2}{*}{HO} & W/O & \multirow{2}{*}{282/251} & 64.59 $\pm$ 3.26 & 65.61 $\pm$ 5.50 & 67.81 $\pm$ 2.84 & 66.36 $\pm$ 4.77 \\
 &                      & W/  &                          & \textbf{68.68 $\pm$ 2.61} & \textbf{69.51 $\pm$ 2.37} & \textbf{70.97 $\pm$ 3.25} & \textbf{70.18 $\pm$ 2.54} \\
\midrule[0.7pt]

\multirow{2}{*}{GTN \cite{Yun2019}}        
 & \multirow{2}{*}{HO} & W/O & \multirow{2}{*}{282/251} & 62.92 $\pm$ 2.85 & 61.60 $\pm$ 5.43 & 61.79 $\pm$ 3.66 & 62.50 $\pm$ 3.86 \\
 &                      & W/  &                          & \textbf{67.09 $\pm$ 3.00} & \textbf{66.30 $\pm$ 3.21} & \textbf{65.68 $\pm$ 3.57} & \textbf{67.69 $\pm$ 3.07} \\
\midrule[0.7pt]

\multirow{2}{*}{GraphSAGE \cite{Hamilton2017}}  
 & \multirow{2}{*}{HO} & W/O & \multirow{2}{*}{282/251} & 63.87 $\pm$ 5.04 & 59.66 $\pm$ 14.94 & 68.85 $\pm$ 12.04 & 59.72 $\pm$ 12.04 \\
 &                      & W/  &                          & \textbf{65.76 $\pm$ 4.67} & \textbf{64.80 $\pm$ 8.26} & \textbf{72.61 $\pm$ 5.98} & \textbf{64.24 $\pm$ 6.66} \\
\bottomrule[1pt]
\end{tabular*}
\end{table*}

\subsection{Robustness and Generalizability}

To further evaluate the robustness and generalizability of the proposed fMRI-Diffusion framework, we investigated whether the synthetic fMRI data generated by our method can consistently improve MDD diagnosis performance across different brain parcellation schemes. This experiment aims to examine whether the benefit of using our synthetic data is independent of the specific atlas used for FC construction.

We adopted six widely used brain atlases that differ in their granularity and parcellation principles: 
(1) the AAL ($R=116$) atlas \cite{Rolls2020}, which partitions the brain based on anatomical landmarks; 
(2) the HO ($R=112$) atlas \cite{Kennedy1998}, which integrates both structural and functional information; 
(3) the Craddock ($R=200$) atlas \cite{Craddock2012}, derived from functional clustering of resting-state correlations; 
(4) the Zalesky ($R=980$) atlas \cite{Zalesky2010}, offering fine-grained functional segmentation; 
(5) the Power ($R=264$) atlas \cite{Power2012}, emphasizing canonical brain networks such as the default mode and sensorimotor systems; and 
(6) the Dosenbach ($R=160$) atlas \cite{Dosenbach2010}, which focuses on regions implicated in cognitive control and attention.  

To ensure a fair evaluation across parcellation schemes, we employed the same MDD classification model, ChebyNet \cite{Defferrard2016}, as the diagnostic backbone. ChebyNet is a spectral graph convolutional network that models higher-order graph structures using Chebyshev polynomial filters, effectively capturing complex functional dependencies across brain regions. All training configurations followed the baseline setup in \cite{Lee2024}, including two graph convolutional layers ($256$ channels, filter size $= 2$), two fully connected layers ($1000$ and $2$ units), the Mish activation function, batch normalization, and the AdamW optimizer with a learning rate of $1\times10^{-5}$ and weight decay of $1\times10^{-3}$. The model was trained for $300$ epochs with a batch size of $32$.

Table \ref{tab:atlas_comparison} compares the diagnostic performance with and without using our synthetic data across the six atlases. As shown, our synthetic data consistently improves performance across all evaluation metrics and all parcellation schemes. Notably, the most substantial gains are observed on finer-grained atlases such as Craddock and Zalesky, indicating that our synthetic fMRI data effectively preserves functional consistency even under complex parcellation. These results further confirm that the proposed framework is robust to varying brain region definitions and can reliably enhance MDD diagnosis regardless of the atlas employed.

\begin{table*}[width=\textwidth,cols=8,pos=t]
\caption{Performance comparison of MDD diagnosis under different brain parcellation atlases, with and without using synthetic fMRI data generated by our proposed method. Results are evaluated using ACC, SEN, SPEC, and F1 (mean~$\pm$~standard deviation), and the better-performing results are highlighted in bold.}
\label{tab:atlas_comparison}

\renewcommand{\arraystretch}{1.1}
\setlength{\tabcolsep}{3pt}
\small
\centering

\begin{tabular*}{\tblwidth}{@{\extracolsep{\fill}} l l c c c c c c @{}}
\toprule[1pt]
\textbf{Method} &
\textbf{Atlas} &
\begin{tabular}[c]{@{}c@{}}\textbf{With/Without}\\\textbf{Our Synth.\ Data}\end{tabular} &
\begin{tabular}[c]{@{}c@{}}\textbf{Samples}\\\textbf{(MDD/NC)}\end{tabular} &
\textbf{ACC (\%)} &
\textbf{SEN (\%)} &
\textbf{SPEC (\%)} &
\textbf{F1 (\%)} \\
\midrule[1pt]

\multirow{12}{*}{ChebyNet \cite{Defferrard2016}} &
\multirow{2}{*}{AAL} &
W/O & \multirow{2}{*}{282/251} &
63.19$\pm$2.01 & 60.95$\pm$4.54 & 65.40$\pm$1.67 & 64.26$\pm$3.20 \\
& &
W/  &  &
\textbf{69.80$\pm$2.81} & \textbf{68.95$\pm$3.34} & \textbf{71.80$\pm$3.86} & \textbf{72.00$\pm$2.95} \\
\cmidrule(lr){2-8}

& \multirow{2}{*}{HO} &
W/O & \multirow{2}{*}{282/251} &
63.00$\pm$3.35 & 62.55$\pm$6.50 & 65.89$\pm$5.20 & 64.35$\pm$4.26 \\
& &
W/  &  &
\textbf{69.25$\pm$2.15} & \textbf{68.95$\pm$3.30} & \textbf{67.90$\pm$3.82} & \textbf{69.90$\pm$2.41} \\
\cmidrule(lr){2-8}

& \multirow{2}{*}{Craddock} &
W/O & \multirow{2}{*}{282/251} &
64.36$\pm$4.24 & 62.52$\pm$8.50 & 64.30$\pm$6.25 & 63.03$\pm$5.88 \\
& &
W/  &  &
\textbf{70.17$\pm$3.25} & \textbf{69.67$\pm$3.23} & \textbf{71.67$\pm$3.53} & \textbf{71.56$\pm$3.15} \\
\cmidrule(lr){2-8}

& \multirow{2}{*}{Zalesky} &
W/O & \multirow{2}{*}{282/251} &
61.36$\pm$4.24 & 62.18$\pm$7.39 & 59.96$\pm$5.08 & 62.08$\pm$5.10 \\
& &
W/  &  &
\textbf{66.35$\pm$1.98} & \textbf{65.66$\pm$4.34} & \textbf{64.49$\pm$1.96} & \textbf{67.51$\pm$2.14} \\
\cmidrule(lr){2-8}

& \multirow{2}{*}{Power} &
W/O & \multirow{2}{*}{282/251} &
59.47$\pm$2.59 & 58.28$\pm$5.15 & 58.66$\pm$9.25 & 58.33$\pm$1.78 \\
& &
W/  &  &
\textbf{62.61$\pm$2.89} & \textbf{61.28$\pm$3.32} & \textbf{66.49$\pm$4.10} & \textbf{65.51$\pm$2.55} \\
\cmidrule(lr){2-8}

& \multirow{2}{*}{Dosenbach} &
W/O & \multirow{2}{*}{282/251} &
57.87$\pm$4.05 & 58.33$\pm$10.24 & 56.44$\pm$6.05 & 59.29$\pm$6.90 \\
& &
W/  &  &
\textbf{61.30$\pm$4.36} & \textbf{62.17$\pm$5.67} & \textbf{61.12$\pm$3.76} & \textbf{62.52$\pm$4.67} \\
\bottomrule[1pt]
\end{tabular*}
\end{table*}

\subsection{Fidelity of Synthetic fMRI Data}

To rigorously assess the fidelity of the synthetic fMRI data generated by our method, we conducted both quantitative and qualitative evaluations. The quantitative assessment focuses on measuring the distributional similarity between real and synthetic data, while the qualitative analysis visually inspects their structural consistency in a reduced-dimensional space.

\subsubsection{Quantitative evaluation}

We employed three complementary statistical metrics to evaluate the similarity between the distributions of real and synthetic fMRI data: the Kullback-Leibler (KL) Divergence, the Wasserstein Distance (WD), and the Kolmogorov-Smirnov (KS) Statistic. These metrics jointly quantify how closely the generated data approximate the real data distribution from different perspectives: information divergence (KL), geometric distance (WD), and cumulative distribution deviation (KS). Together, they provide a comprehensive and robust assessment of distributional fidelity, where lower values indicate higher similarity between the two datasets.

Table \ref{tab:distribution} summarizes the quantitative comparison results. Across both MDD and NC classes, the KL, WD, and KS values remain consistently low (all below 0.06), confirming that the synthetic samples closely follow the real data distribution. These results validate the statistical fidelity of the generated data and demonstrate that the proposed fMRI-Diffusion framework effectively preserves the intrinsic characteristics of the original fMRI signals.
\begin{table}[htbp]
\caption{Quantitative evaluation of the distributional similarity between real and synthetic fMRI data using KL divergence, WD, and KS statistic. Lower values indicate higher similarity.}
\centering
\label{tab:distribution}
\setlength{\tabcolsep}{3pt}
\begin{tabular}{cccc}
\toprule[1pt]
\textbf{Class} & \textbf{KL Divergence} & \textbf{Wasserstein Distance} & \textbf{KS Statistic} \\
\midrule[1pt]
MDD & 0.0183 & 0.0479 & 0.0522 \\

NC  & 0.0181 & 0.0471 & 0.0511 \\
\bottomrule[1pt]
\end{tabular}
\end{table}

\subsubsection{Qualitative evaluation}

\begin{figure*}
	\centering
	\includegraphics[width=\linewidth]{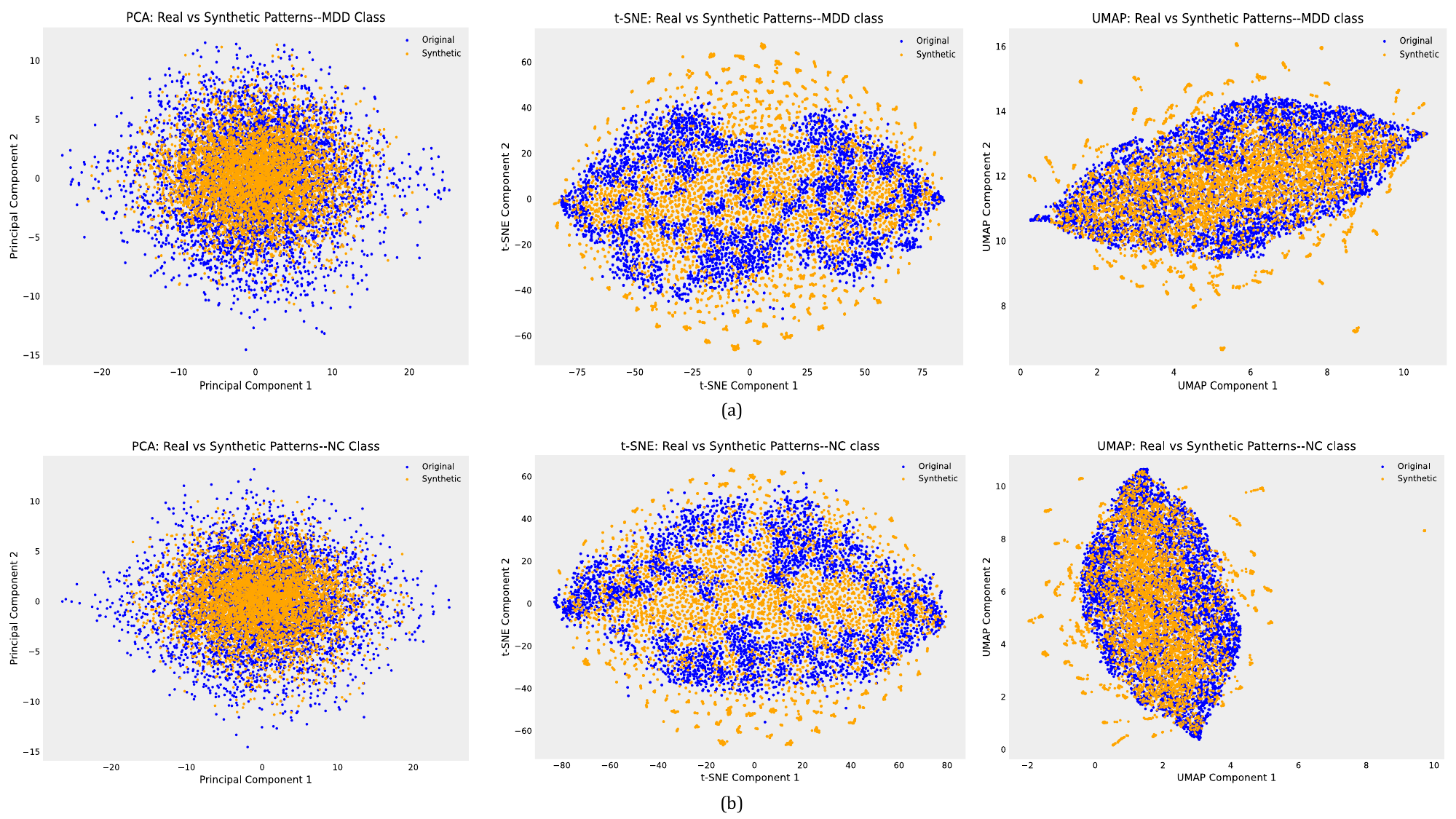}
	\caption{PCA, t-SNE, and UMAP visualizations of real and synthetic fMRI data generated by the proposed fMRI-Diffusion model for the MDD and NC classes. Each point corresponds to a single time point (ROI activation pattern). Blue and yellow points denote real and synthetic data, respectively. The substantial overlap between the two distributions across all three embedding methods indicates that the proposed model effectively captures the underlying activation patterns and preserves both the global variance structure and local structure observed in the real fMRI data.}
	\label{fig:combined}
\end{figure*}

Beyond numerical analysis, we further examined the structural correspondence between real and synthetic samples using principal component analysis (PCA), t-distributed Stochastic Neighbor Embedding (t-SNE), and Uniform Manifold Approximation and Projection (UMAP). The two-dimensional projections obtained from PCA, t-SNE, and UMAP shown in Fig. \ref{fig:combined}, reveals a high degree of overlap between the distributions of real (blue) and synthetic (yellow) samples across both MDD and NC groups. This overlap indicates that the synthetic data accurately captures the major variance components and structural patterns inherent in the real fMRI data. Moreover, the synthetic samples uniformly populate the manifold of the original data without mode collapse or over-dispersion, confirming that our diffusion-based generation process maintains appropriate diversity while preserving class-specific distributional boundaries. Together, these findings demonstrate that the proposed framework generates realistic, statistically consistent, and structurally faithful fMRI time series suitable for downstream diagnostic applications.

\subsection{Superiority of fMRI Time Series Synthesis over FC Synthesis}

To further demonstrate the advantage of synthesizing fMRI time series over FC matrices, we conducted a controlled comparison using the same generative framework and classification backbone. In this experiment, the proposed model was separately trained to synthesize (1) FC matrices and (2) fMRI time series. The FC-based synthesis was performed by feeding FC matrices directly into the diffusion model, whereas the fMRI-based synthesis followed our standard framework that models ROI-level temporal dynamics before FC computation.

For evaluation, the same number of synthetic samples and the same ChebyNet classifier \cite{Defferrard2016} were employed across both settings to ensure a fair comparison. The classification performance was assessed under six brain parcellation atlases, using identical training and testing protocols as in previous experiments. As reported in Table \ref{tab:fMRI_vs_FC_comparison}, the diagnostic performance obtained using synthetic fMRI data consistently surpasses that of synthetic FC data across all atlases and evaluation metrics. The improvements are particularly notable for the AAL, HO, and Craddock atlases, with accuracy gains of up to 2 to 3 percentage points. These results clearly indicate that generating fMRI time series preserves richer temporal and structural dependencies in brain activity than directly synthesizing static FC representations. Consequently, fMRI-based synthesis provides more informative and biologically meaningful data for downstream MDD diagnosis.

\begin{table*}[width=\textwidth,cols=8,pos=htbp]
\caption{Comparison of MDD diagnosis performance using synthetic fMRI time series and synthetic FC data generated by the proposed method across different brain atlases. Results are evaluated in terms of ACC, SEN, SPEC, and F1, and reported as mean~$\pm$~standard deviation. The better-performing results are highlighted in bold.}
\label{tab:fMRI_vs_FC_comparison}

\renewcommand{\arraystretch}{1.1}
\centering
\setlength{\tabcolsep}{3pt}
\small

\begin{tabular*}{\tblwidth}{@{\extracolsep{\fill}} l c l c c c c c @{}}
\toprule[1pt]
\textbf{Method} &
\begin{tabular}[c]{@{}c@{}}\textbf{Use Synth.}\\\textbf{fMRI / FC}\end{tabular} &
\textbf{Atlas} &
\begin{tabular}[c]{@{}c@{}}\textbf{Samples}\\\textbf{(MDD/NC)}\end{tabular} &
\textbf{ACC (\%)} &
\textbf{SEN (\%)} &
\textbf{SPEC (\%)} &
\textbf{F1 (\%)} \\
\midrule[1pt]

\multirow{12}{*}{ChebyNet \cite{Defferrard2016}} &
FC   & \multirow{2}{*}{AAL}       & \multirow{2}{*}{282/251} & 68.75$\pm$3.00 & 67.50$\pm$3.90 & 70.90$\pm$3.90 & 71.56$\pm$3.15 \\
& fMRI &                            &                          & \textbf{69.80$\pm$2.81} & \textbf{68.95$\pm$3.34} & \textbf{71.80$\pm$3.86} & \textbf{72.00$\pm$2.95} \\
\cmidrule(lr){2-8}

& FC   & \multirow{2}{*}{HO}       & \multirow{2}{*}{282/251} & 67.30$\pm$2.25 & 66.80$\pm$3.70 & 66.00$\pm$4.00 & 67.90$\pm$2.65 \\
& fMRI &                            &                          & \textbf{68.00$\pm$2.15} & \textbf{67.95$\pm$3.30} & \textbf{66.79$\pm$3.82} & \textbf{68.37$\pm$2.41} \\
\cmidrule(lr){2-8}

& FC   & \multirow{2}{*}{Craddock} & \multirow{2}{*}{282/251} & 70.00$\pm$3.10 & 68.67$\pm$3.20 & 71.60$\pm$3.60 & 71.20$\pm$3.30 \\
& fMRI &                            &                          & \textbf{70.17$\pm$3.25} & \textbf{69.67$\pm$3.23} & \textbf{71.67$\pm$3.53} & \textbf{71.56$\pm$3.15} \\
\cmidrule(lr){2-8}

& FC   & \multirow{2}{*}{Zalesky}  & \multirow{2}{*}{282/251} & \textbf{67.55$\pm$2.90} & \textbf{67.60$\pm$4.55} & \textbf{65.60$\pm$1.88} & 67.50$\pm$2.34 \\
& fMRI &                            &                          & 66.35$\pm$1.98 & 65.66$\pm$4.34 & 64.49$\pm$1.96 & \textbf{67.51$\pm$2.14} \\
\cmidrule(lr){2-8}

& FC   & \multirow{2}{*}{Power}    & \multirow{2}{*}{282/251} & 61.55$\pm$3.10 & \textbf{61.38$\pm$3.50} & 66.20$\pm$4.60 & 64.95$\pm$2.90 \\
& fMRI &                            &                          & \textbf{62.61$\pm$2.89} & 61.28$\pm$3.32 & \textbf{66.49$\pm$4.10} & \textbf{65.51$\pm$2.55} \\
\cmidrule(lr){2-8}

& FC   & \multirow{2}{*}{Dosenbach} & \multirow{2}{*}{282/251} & 60.80$\pm$4.56 & 61.96$\pm$4.55 & \textbf{61.96$\pm$3.85} & 62.20$\pm$4.75 \\
& fMRI &                             &                          & \textbf{61.30$\pm$4.36} & \textbf{62.17$\pm$5.67} & 61.12$\pm$3.76 & \textbf{62.52$\pm$4.67} \\
\bottomrule[1pt]
\end{tabular*}
\end{table*}

\begin{table*}[width=\textwidth,cols=6,pos=t]
\caption{Ablation study of the proposed fMRI-Diffusion framework. Results are reported as mean $\pm$ standard deviation over 5-fold cross-validation with five random seeds. In each fold, pretraining, diffusion training, and synthetic data generation are performed using the training data only.}
\label{tab:ablation}

\renewcommand{\arraystretch}{1.1}
\setlength{\tabcolsep}{4pt}
\centering
\small

\begin{tabular*}{\tblwidth}{@{\extracolsep{\fill}} p{6.2cm} c c c c c @{}}
\toprule[1pt]
\textbf{Variant} &
\begin{tabular}[c]{@{}c@{}}\textbf{With/Without}\\\textbf{Our Synth.\ Data}\end{tabular} &
\textbf{ACC (\%)} &
\textbf{SEN (\%)} &
\textbf{SPEC (\%)} &
\textbf{F1 (\%)} \\
\midrule[1pt]

\multirow{2}{*}{\parbox[t]{6.2cm}{\textbf{Full model}\\(Supervised pretraining + Transformer denoiser)}}
& W/O & 64.36$\pm$4.24 & 62.52$\pm$8.50 & 64.30$\pm$6.25 & 63.03$\pm$5.88 \\
& W/  & \textbf{70.17$\pm$3.25} & \textbf{69.67$\pm$3.23} & \textbf{71.67$\pm$3.53} & \textbf{71.26$\pm$3.15} \\[0.8ex]
\cmidrule(lr){2-6}

\multirow{2}{*}{\parbox[t]{6.2cm}{Without pretraining\\(Transformer denoiser, random initialization)}}
& W/O & 64.36$\pm$4.24 & 62.52$\pm$8.50 & 64.30$\pm$6.25 & 63.03$\pm$5.88 \\
& W/  & \underline{68.25$\pm$2.15} & \underline{67.96$\pm$3.30} & 65.79$\pm$4.59 & \underline{68.57$\pm$3.26} \\[0.8ex]
\cmidrule(lr){2-6}

\multirow{2}{*}{\parbox[t]{6.2cm}{1D U-Net denoiser\\(replacing the Transformer denoiser)}}
& W/O & 64.36$\pm$4.24 & 62.52$\pm$8.50 & 64.30$\pm$6.25 & 63.03$\pm$5.88 \\
& W/  & 66.26$\pm$4.25 & 65.57$\pm$4.26 & \underline{67.68$\pm$3.42} & 67.56$\pm$5.15 \\[0.8ex]
\cmidrule(lr){2-6}

\multirow{2}{*}{\parbox[t]{6.2cm}{LSTM denoiser\\(replacing the Transformer denoiser)}}
& W/O & 64.36$\pm$4.24 & 62.52$\pm$8.50 & 64.30$\pm$6.25 & 63.03$\pm$5.88 \\
& W/  & 67.35$\pm$3.98 & 66.96$\pm$5.34 & 66.49$\pm$3.96 & 65.51$\pm$3.14 \\
\bottomrule[1pt]
\end{tabular*}
\end{table*}

\subsection{Ablation Experiments}

To assess the contribution of the primary design choices in the proposed fMRI-Diffusion framework, we conducted controlled ablation experiments. All settings follow the main experimental protocol. In each fold, pretraining, diffusion training, and synthetic sample generation are performed exclusively on the training set. The data augmentation procedure is held constant across all variants, whereas the denoising network is modified according to the corresponding ablation setting. In all experiments, we use the Craddock atlas ($R=200$) \cite{Craddock2012}.

We evaluate the full model against three variants: (1) training without supervised pretraining (random initialization of the Transformer encoder), (2) replacing the Transformer denoiser with a 1D U-Net, and (3) replacing the Transformer denoiser with an LSTM. For consistency, all variants use the same data split protocol and the same downstream diagnosis backbone, i.e., ChebyNet \cite{Defferrard2016}.

Table~\ref{tab:ablation} summarizes the results. The rows labeled ``W/O'' are identical across all variants because the downstream classifier, real training data, and evaluation protocol are the same when synthetic augmentation is not applied. The effect of each ablation is therefore reflected in the ``W/'' rows, where synthetic samples generated by each variant are added to the training set. Under the augmented setting, the full model (Pretraining + Transformer Encoder) achieves the highest performance across all reported metrics. The comparison between the full model and the variant without pretraining shows the contribution of supervised initialization. Without pretraining, augmentation still improves performance (ACC $68.25 \pm 2.15$), but the full model remains higher in ACC, SEN, SPEC, and F1. This suggests that pretraining helps the denoising model incorporate temporal structure before diffusion training, which is beneficial for reverse process modeling. Replacing the Transformer denoiser with alternative architectures leads to lower augmented performance. The LSTM variant reaches ACC $67.35 \pm 3.98$ and the 1D U-Net variant reaches ACC $66.26 \pm 4.25$, both below the full model. Although both alternatives still improve over the non-augmented baseline, the gap relative to the Transformer-based denoiser indicates that the self-attention mechanism plays an important role in capturing dependencies across the temporal axis of fMRI sequences.

\subsection{Cross-Site Generalization}

To further examine the generalizability of the proposed fMRI-Diffusion framework, we evaluated its performance across multiple independent sites within the REST-meta-MDD dataset, which exhibit notable inter-site variability in scanning conditions and participant demographics. Specifically, data from the three largest contributing sites (S25, S21, and S8) were used, containing 152, 156, and 150 subjects, respectively. All experiments employed the Craddock atlas for consistent parcellation and the ChebyNet classifier \cite{Defferrard2016} for MDD diagnosis.

As summarized in Table \ref{tab:site_comparison}, incorporating synthetic fMRI data generated by our model consistently improved diagnostic performance across all sites and evaluation metrics. For instance, on Site 25, accuracy increased from 60.28\% to 63.65\%; similar gains were observed on Site 21 (from 56.83\% to 57.90\%) and Site 8 (from 64.69\% to 70.55\%). These results confirm that our synthetic data not only enhances within-site diagnosis but also provides stable benefits under distributional shifts between sites. The observed improvements highlight the robustness and generalization capability of the proposed method, supporting its potential applicability to real-world multi-site clinical studies.
\begin{table*}[width=\textwidth,cols=9,pos=ht]
\caption{Cross-site evaluation of MDD diagnosis performance with and without using synthetic fMRI data generated by the proposed method. Performance is reported as mean~$\pm$~standard deviation, with the better-performing results highlighted in bold.}
\label{tab:site_comparison}

\renewcommand{\arraystretch}{1.1}
\centering
\setlength{\tabcolsep}{3pt}
\small

\begin{tabular*}{\tblwidth}{@{\extracolsep{\fill}} l l c l c c c c c @{}}
\toprule[1pt]
\textbf{Method} &
\textbf{Site} &
\begin{tabular}[c]{@{}c@{}}\textbf{With/Without}\\\textbf{Our Synth.\ Data}\end{tabular} &
\textbf{Atlas} &
\begin{tabular}[c]{@{}c@{}}\textbf{Samples}\\\textbf{(MDD/NC)}\end{tabular} &
\textbf{ACC (\%)} &
\textbf{SEN (\%)} &
\textbf{SPEC (\%)} &
\textbf{F1 (\%)} \\
\midrule[1pt]

& \multirow{2}{*}{S25} &
W/O &
\multirow{2}{*}{Craddock} &
\multirow{2}{*}{89/63} &
60.28$\pm$10.35 &
53.72$\pm$19.14 &
66.92$\pm$7.37 &
56.24$\pm$15.27 \\
& &
W/  &
 &
 &
\textbf{63.65$\pm$2.10} &
\textbf{59.69$\pm$14.10} &
\textbf{69.62$\pm$10.17} &
\textbf{58.98$\pm$4.25} \\
\cmidrule(lr){2-9}

\multirow{2}{*}{ChebyNet \cite{Defferrard2016}} &
\multirow{2}{*}{S21} &
W/O &
\multirow{2}{*}{Craddock} &
\multirow{2}{*}{86/70} &
56.83$\pm$9.81 &
56.37$\pm$14.46 &
57.14$\pm$8.75 &
55.99$\pm$11.39 \\
& &
W/  &
 &
 &
\textbf{57.90$\pm$8.40} &
\textbf{59.13$\pm$6.52} &
\textbf{60.10$\pm$10.10} &
\textbf{58.00$\pm$6.93} \\
\cmidrule(lr){2-9}

& \multirow{2}{*}{S8} &
W/O &
\multirow{2}{*}{Craddock} &
\multirow{2}{*}{75/75} &
64.69$\pm$8.05 &
61.18$\pm$13.49 &
67.71$\pm$11.58 &
61.76$\pm$10.58 \\
& &
W/  &
 &
 &
\textbf{70.55$\pm$4.56} &
\textbf{67.20$\pm$6.84} &
\textbf{72.56$\pm$6.61} &
\textbf{68.00$\pm$5.16} \\
\bottomrule[1pt]
\end{tabular*}
\end{table*}

\section{Discussion}

In this study, we presented an fMRI time-series synthesis framework for the diagnostic assessment of MDD based on a denoising diffusion probabilistic model with a temporal Transformer denoiser. The framework was developed to address two practical challenges in clinical fMRI studies of MDD, namely limited sample size and the loss of temporal information in functional-connectivity-based synthesis. By modeling the fMRI signal directly in the time domain, the proposed fMRI-Diffusion framework preserves temporal dynamics and learns a denoising function that captures long-range dependencies in ROI time series.

Limited data availability remains an important issue for MDD classification. Although public datasets such as REST-meta-MDD provide a valuable resource, effective use of multi-site data is challenging because scanner differences, acquisition settings, and cohort characteristics introduce site-related distribution variation. In this setting, the temporal Transformer provides a flexible sequence modeling component for the reverse diffusion process, allowing the model to learn temporal structure from the training data and improve the quality of synthesized sequences. When synthetic samples generated by the proposed framework are combined with the original training set, the results in Tables~\ref{tab:mdd_comparison} and~\ref{tab:merged_performance_corrected} show improved diagnostic classification performance in several settings. In addition, the site-wise results in Table~\ref{tab:site_comparison} indicate that the framework maintains favorable performance across independent sites within the REST-meta-MDD dataset, despite inter-site variation in acquisition protocols and participant characteristics. The ablation results further clarify the contribution of supervised pretraining to the overall framework. As shown in Table~\ref{tab:ablation}, the pretraining stage improves the performance of the Transformer-based denoiser when synthetic data are used for augmentation. A plausible explanation is that supervised pretraining initializes the Transformer encoder with task-relevant temporal representations before diffusion training, which supports more stable learning of the reverse diffusion process. This initialization helps the denoiser capture temporal patterns in fMRI signals more effectively, leading to synthetic samples that are more useful for downstream classification. The improvement observed in the augmented setting suggests that pretraining contributes not only to sequence modeling quality, but also to the diagnostic value of the generated data.

Recent studies have explored generative modeling of fMRI signals in different settings, and the present work contributes to this line of research by focusing on diffusion-based synthesis of fMRI time-series data for MDD diagnosis and by examining its utility for data augmentation in a multi-site clinical dataset. Data scarcity is also a concern for other fMRI-based disorders, including attention deficit hyperactivity disorder, autism spectrum disorder, mild cognitive impairment, and Parkinson's disease. The proposed framework may therefore provide a useful basis for future methods designed to address limited sample size in these conditions. A natural extension of this work is to evaluate fMRI-Diffusion on additional neuropsychiatric and neurological disorders and to study whether the same synthesis strategy provides similar benefits across different diagnostic tasks. In addition, multimodal fusion of imaging and non-imaging data may offer complementary information and a more comprehensive description of brain function and network organization~\cite{cai2025mm}. Future research may extend the current framework to multimodal data synthesis, with the goal of modeling both intra-modal temporal structure and inter-modal relationships. Such extensions may further improve diagnostic performance for MDD and may also support broader applications in neuropsychiatric disorder analysis.

\section{Conclusion}

In this work, we introduced \textbf{fMRI-Diffusion}, a novel diffusion-based generative framework designed to synthesize realistic fMRI time series for MDD diagnosis. Unlike methods that synthesize FC matrices, our framework generates fMRI time series at the ROI level using a pretrained Transformer-based denoiser, enabling fine-grained reconstruction of brain activation patterns. The proposed framework effectively captures the complex temporal dependencies inherent in resting-state fMRI signals while maintaining structural fidelity and biological plausibility. Extensive experiments on the REST-meta-MDD dataset demonstrate the superior performance of our method across multiple evaluation dimensions. The synthesized fMRI data substantially improve the accuracy and robustness of various MDD diagnostic models, enhance performance across different brain parcellations, and generalize well across independent data collection sites. Quantitative and qualitative analyses further confirm the high fidelity and distributional consistency between synthetic and real fMRI data, validating the reliability of the proposed generative model. Overall, this work introduces a diffusion-based approach for fMRI time series synthesis that complements FC-based augmentation methods in neuroimaging-based MDD diagnosis. Future work will explore extending the framework to multimodal neuroimaging data and cross-dataset generalization to further enhance clinical applicability and diagnostic precision.

\printcredits

\bibliographystyle{unsrtnat}
\bibliography{cas-refs}

\end{document}